\documentclass[letterpaper, 10 pt, conference]{ieeeconf}  

\usepackage{kotex}
\usepackage[utf8]{inputenc}
\usepackage[T1]{fontenc}
\usepackage{lmodern}
\usepackage{graphicx}
\usepackage{amsmath,mathtools}
\let\labelindent\relax
\usepackage{enumitem}
\usepackage{ragged2e} 
\usepackage{microtype} 
\usepackage{booktabs}
\usepackage{makecell}   
\usepackage{dashrule}   
\usepackage{balance}
\IEEEoverridecommandlockouts                              

\title{\LARGE \bf
Contrast-Guided Cross-Modal Distillation for Thermal Object Detection
}

\author{SiWoo Kim$^{1}$ and JhongHyun An$^{1}$
\thanks{$^{1}$Gachon University, Seongnam, Republic of Korea
        {\tt\small ksw5830@gachon.ac.kr}}%
\thanks{$^{1}$Gachon University, Seongnam, Republic of Korea
        {\tt\small jhonghyun@gachon.ac.kr}}%
\thanks{Corresponding author: Jhonghyun An}%
}

\begin{document}

\maketitle
\thispagestyle{empty}
\pagestyle{empty}

\begin{abstract}

Robust perception at night remains challenging for thermal-infrared detection: low contrast and weak high-frequency cues lead to duplicate, overlapping boxes, missed small objects, and class confusion. Prior remedies either translate TIR to RGB and hope pixel fidelity transfers to detection—making performance fragile to color, structure artifacts—or fuse RGB and TIR at test time, which requires extra sensors, precise calibration, and higher runtime cost. Both lines can help in favorable conditions, but do not directly shape the thermal representation used by the detector.
We keep mono-modality inference and tackle the root causes during training. Specifically, we introduce training-only objectives that sharpen instance-level decision boundaries by pulling together features of the same class and pushing apart those of different classes—suppressing duplicate and confusing detections—and that inject cross-modal semantic priors by aligning the student’s multi-level pyramid features with an RGB-trained teacher, thereby strengthening texture-poor thermal features without visible input at test time. In experiments, our method outperformed prior approaches and achieved state-of-the-art performance.

\end{abstract}

\par\medskip
\section{INTRODUCTION}

Autonomous driving has advanced rapidly with deep learning, yet robust perception remains challenging under adverse illumination and weather. Practical systems employ multiple sensors—RGB cameras, LiDAR, and radar—but each modality degrades under low light or inclement conditions. Cost considerations further push practitioners toward camera-centric setups, which are especially vulnerable at night due to the absence of visible illumination.

\setlength{\parindent}{0.1in} Thermal infrared(TIR) cameras are widely used in autonomous driving, surveillance, and defense because they sense emitted heat and thus enable detection at night. However, effective detection requires a temperature contrast between objects and background. Compared with RGB, TIR images typically lack color and often exhibit weaker edge and high-frequency texture cues, which can depress classification and detection performance.

\setlength{\parindent}{0.1in} To compensate for missing color and fine structure, prior work has explored translating TIR to visible images. As shown in Fig. 1 (a), Early approaches leveraged Generative Adversarial Network(GAN) \cite{goodfellow2014gan}, trained with a generator that synthesizes images and a discriminator that distinguishes real from synthetic. The generator’s objective is to produce images realistic enough to fool the discriminator, while the discriminator learns to correctly identify fakes. Beyond pure synthesis, GANs have been extended to cross-domain translation: Pix2Pix \cite{isola2017pix2pix} conditions the generator on the input and optimizes an adversarial objective plus a pixelwise L1 loss against the paired ground truth; CycleGAN \cite{zhu2017cyclegan} introduces cycle-consistency to enable translation without aligned pairs.

\setlength{\parindent}{0.1in} As illustrated in Fig. 1 (b), rather than relying solely on translated imagery, many systems fuse RGB and TIR to exploit complementary strengths—RGB color, texture with TIR thermal cues. Fusion is commonly categorized as: early fusion(concatenating raw modalities at input), mid-level fusion(merging intermediate backbone features), and late fusion (combining decisions or high-level representations). While effective, fusion incurs practical burdens, including precise cross-sensor calibration and increased hardware cost.

\begin{figure}[t]
  \centering
  \includegraphics[width=\linewidth,height=.88\textheight,keepaspectratio]{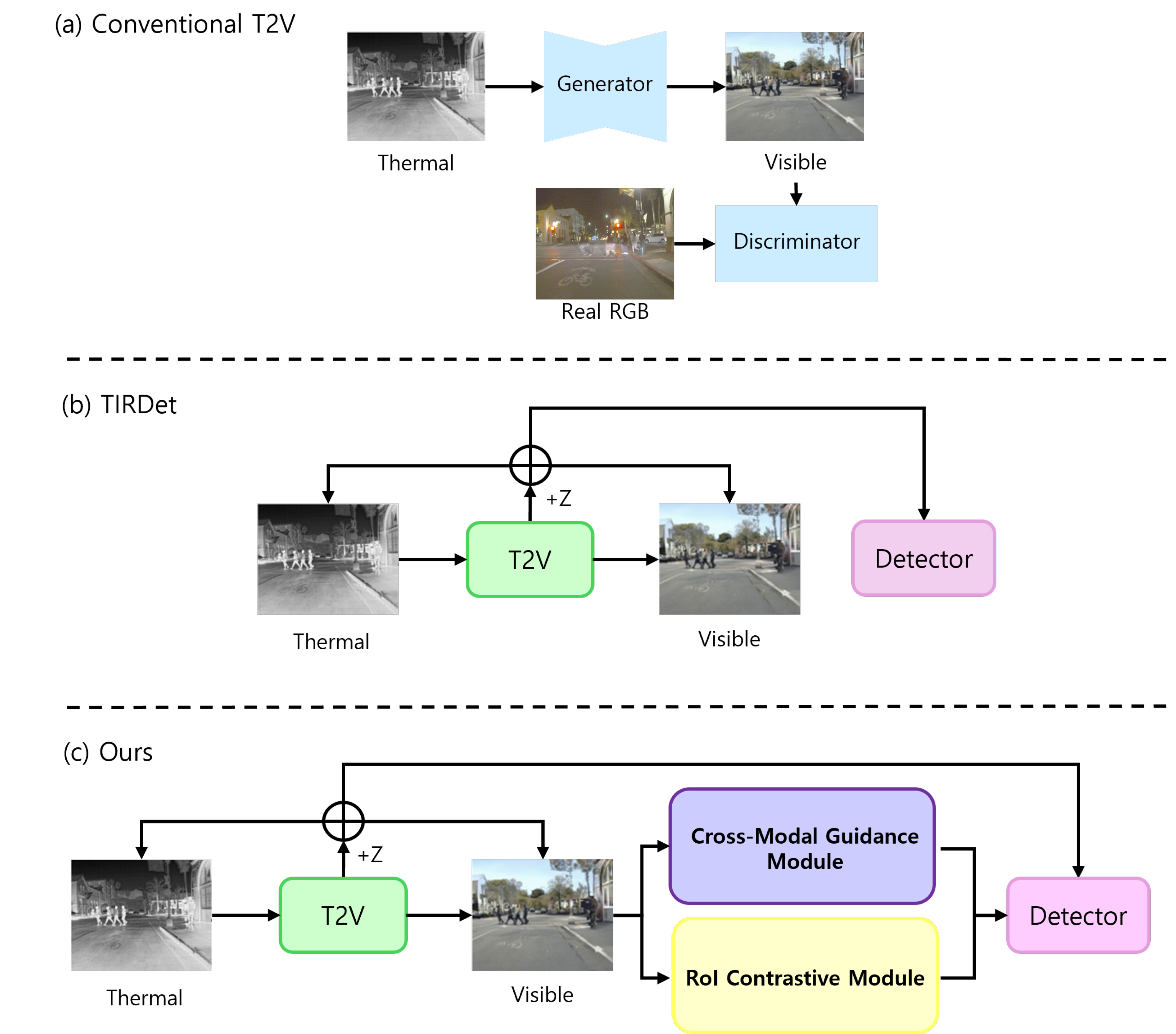}
  \vspace{-0.7cm}
  \caption{\textbf{(a) Conventional thermal-to-visible(T2V).} A pix2pix \cite{isola2017pix2pix}-style translator with a U-Net generator and a PatchGAN discriminator.  \textbf{(b) TIRDet \cite{wang2023tirdet}.} A frozen T2V produces a pseudo-RGB that is early-fused with thermal. \textbf{(c) Ours}. The RCS module improves class separability in the feature space.}
  \vspace{-2mm}
  \label{fig:tirdet}
\end{figure}

\setlength{\parindent}{0.1in} A related line of research synthesizes a cross-domain image at the same timestamp(e.g., generate RGB from TIR) and then fuses the pair for detection. These pipelines remain sensitive to translation quality: artifacts, color shifts, or structural inconsistencies between the generated RGB and the original TIR can propagate to downstream detection.

\begin{figure*}[t]
  \centering
  \includegraphics[width=\textwidth,trim=0 0 0 0,clip]{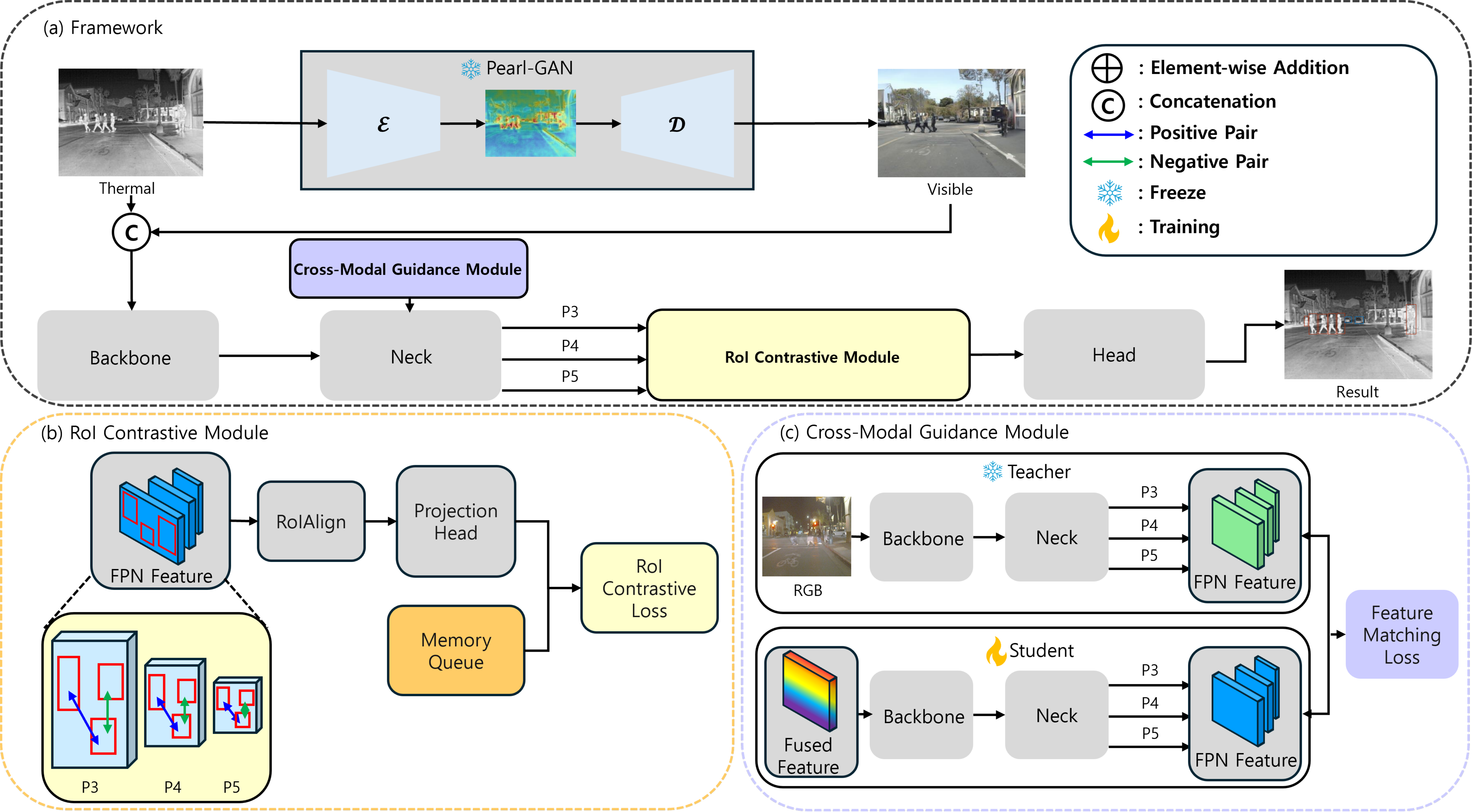}
  \vspace{-0.7cm}
  \caption{\textbf{Overview of the Framework} {(a) The framework constructs with RoI Contrastive(RCS) Module, Cross-Modal Guidance(CMG) Module} {(b) RCS applies supervised contrastive learning to GT-aligned RoI embeddings to enforce a class-separable space.} {(c) CMG aligns teacher RGB features and student thermal RoI features via feature-level distillation; inference uses thermal only.}}
  \vspace{-0.3cm}
  \label{fig:tirdet}
\end{figure*}

\setlength{\parindent}{0.1in} To address these limitations, we propose CGDet. As depicted in Fig. 1 (c), our method increases representational separability during training via RoI-based contrastive learning and injects semantics through multi-level feature alignment with an RGB teacher, while inference uses a thermal-only input and incurs no additional overhead.
\setlist[itemize]{leftmargin=*,align=left} 
\begin{itemize}[leftmargin=*,itemsep=0.25ex,topsep=0.25ex]
  \item {We propose RCS}, a RCS that pulls same-class RoI features together and pushes different-class features apart, reducing false positives under low contrast.
  \item {We introduce CMG}, a feature-level cross-modal guidance module that distills semantics from an RGB-trained teacher into the thermal FPN \cite{lin2017fpn} during training, with no extra sensors or overhead at inference.
  \item {We keep practicality}, matching the base detector’s inference cost while improving discriminability.
  \item {Extensive experiments} on FLIR \cite{flir} show consistent gains in the mono-modality setting and competitiveness with multispectral(V+T) methods.
\end{itemize}

\par\medskip
\section{RELATED WORK}

\subsection{Thermal-to-Visible Translation}

\setlength{\parindent}{0.1in} Thermal-to-Visible(T2V) is more challenging than generic image-to-image(I2I) due to cross-spectral gaps(radiometry, emissivity, wavelength bands,illumination dependence) and frequent fine-scale misalignment between TIR and RGB. Paired T2V(e.g., TIC-CGAN \cite{kuang2020tic-cgan}, Pearl-GAN \cite{luo2022pearl-gan}) typically couples adversarial learning with L1/L2, perceptual, TV, and edge gradient terms, stabilizing color statistics and preserving boundaries; when pairs are tightly aligned, these methods achieve high structural fidelity and connect cleanly to downstream detectors, but even minor misalignment can induce ghosting and boundary artifacts. Unpaired T2V(e.g., IR2VI \cite{liu2018ir2vi}) maintains cross-domain correspondence without pairs using cycle, identity constraints and augments them with structure connections and RoI weighting to emphasize object contours and fine details. While unpaired training increases data flexibility, it is sensitive to loss balancing and can still exhibit cross-spectral correspondence errors.

\subsection{Contrastive Learning}

\setlength{\parindent}{0.1in} Contrastive learning structures the feature space by pulling semantically similar samples together and pushing dissimilar ones apart; unsupervised variants(e.g., SimCLR \cite{chen2020simclr}, MoCo \cite{he2020moco}) use augmented positives, while SupCon \cite{khosla2020supcon} leverages labels to boost class separability. In detection, contrastive losses complement standard cls, box objectives at image-, proposal-, or instance-level, yet most studies target RGB settings where features are strong. Thermal detection, however, suffers from low contrast and weak textures, yielding overlapping class representations. 

\setlength{\parindent}{0.1in} We introduce the RCS: GT-aligned RoIs are sampled from the feature pyramid and optimized with a supervised contrastive loss to enforce global class separability, complementing losses and improving robustness under thermal conditions.

\subsection{Knowledge distillation}
\setlength{\parindent}{0.1in} Knowledge distillation transfers the knowledge of a large teacher model to a lightweight student, improving the accuracy–efficiency trade-off. Early work \cite{hinton2015distil} began with response-based distillation in which the student mimics the teacher’s softened logits. Subsequent studies expanded to feature-based distillation that aligns intermediate representations; notable examples include FitNets \cite{romero2015fitnet}, which uses shallow intermediate layers as “hints” to guide the student, and Attention Transfer \cite{zagoruyko2016attention_distill}, which matches activation, attention maps. More recently, contrastive learning has been incorporated—CRD \cite{tian2019contrastive_disitll} aligns teacher–student features by pulling positives together and pushing negatives apart. For dense prediction tasks such as object detection, methods that distill multi-level FPN features, RoI, anchor-level local features, and relational, attention cues—e.g., G-DetKD \cite{yao2021G-detkd}—are widely adopted. In cross-modal settings (e.g., RGB–thermal), cross-modal/cross-domain distillation transfers semantic cues from an RGB teacher into the feature space of a thermal student to enhance mono-modality inference; representative approaches include CrossModalKD \cite{feng2023cekd}. In this context, our CMG injects semantic information from an RGB teacher into the student’s multi-level FPN features, while RCS complements it by increasing the global separability of RoI representations via a contrastive separation objective. Both modules are active only during training and add no inference overhead, yet they deliver improved accuracy using only thermal input.

\par\medskip
\section{METHOD}

\subsection{Overview}
\textbf{Overall architecture}. 
We aim to enhance thermal object detection with RGB-guided training while keeping inference thermal-only. We add two RoI-based auxiliary losses used only during training: RoI Contrastive module(RCS) and Cross-Modal Guidance module. Fig 2 shows the framework. The detector ingests thermal input; during training, it may also fuse generated RGB. We sample GT-aligned RoIs and enforce supervised contrast for class-separable embeddings and feature distillation from an RGB-trained teacher. In inference, the teacher and any generated RGB are removed, so the model incurs no extra cost.

\subsection{RoI-level Supervised Contrastive Learning} 
Thermal images often suffer from low contrast and limited resolution, which leads to representation overlap between instances of different classes in the feature space, resulting in similar local patterns. Conventional one-stage detectors such as the YOLOX \cite{ge2021yolox} family rely solely on per-location classification and regression losses, and thus do not explicitly enforce global class separability in the representation space. 

Thermal images often have low contrast and limited detail. Different classes can overlap in the representation space. One-stage detectors optimize local classification and regression but do not enforce global class separability. We introduce RCS. During training, we sample GT-aligned RoIs from the FPN. We compute a supervised contrastive loss that pulls same-class embeddings together and pushes different-class embeddings apart. This reduces confusion under low contrast and improves robustness.

\setlength{\parindent}{0.1in} To address this issue, we propose RCS. During training, ground-truth(GT) bounding boxes are used to sample RoI-aligned features from the feature pyramid. A supervised contrastive loss is then imposed such that features from the same class are pulled closer together, while those from different classes are pushed further apart. Each GT box is assigned to an FPN-level based on its size, and from the selected level’s feature map, an \( l \times l \) patch is extracted via RoIAlign \cite{he2017mask-rcnn}. The extracted RoI feature is subsequently transformed into an embedding vector \( z \) through Global Average Pooling(GAP) followed by a linear projection layer. Given \( M \) RoI embeddings and their class labels in a mini-batch, the contrastive loss for a sample \( i \) with temperature \( \tau \) is defined as:

\begin{equation}
\mathcal{L}_i = -\frac{1}{|P(i)|} 
\sum_{p \in P(i)} 
\log \frac{\exp \left( z_i \cdot z_p / \tau \right)}
{\sum_{a \in A(i)} \exp \left( z_i \cdot z_a / \tau \right)}
\end{equation}

To enrich the pool of hard negatives, we maintain a memory queue of size \( K \), and concatenate its embeddings with those of the current batch to construct the negative pool. In this way, RCS brings together representations of RoIs from the same class while pushing apart those from different classes, thereby mitigating class confusion induced by low contrast and resolution in thermal imagery and strengthening the global separability of instance-level features.
\par\medskip

\subsection{Feature-level Cross-Modal Supervision} 
Thermal features often lack rich semantic cues, making it difficult for the detector to learn discriminative representations. To mitigate this, we introduce a CMG that leverages a frozen teacher network trained on RGB to guide the thermal-based student. The teacher is a backbone–neck network pre-trained on RGB and frozen during training. The student takes as input the concatenation of thermal and visible channels, producing multi-level FPN features. Both the teacher and the student produce pyramid features \( {T_k} \) and \( S_k \) at levels $k \in \{3,4,5\}$. To align the student features with the teacher’s semantic knowledge, we enforce a feature-level consistency loss:

\begin{equation}
\label{eq:cms}
\mathcal{L}_{\mathrm{cms}}
= \sum_{k} \alpha_k \bigl(
  \| S_k - T_k \|_{1}
  + \lambda \,[1 - \cos(S_k, T_k)]
\bigr)
\end{equation}

where $\lVert \cdot \rVert_{1}$ is an L1 loss ensuring local similarity, and the cosine term stabilizes directionality in feature space. \( {\alpha_k} \) balances pyramid levels. During training, gradients flow only through the student; the teacher remains frozen. At inference time, only the student path is retained, so the pipeline has no additional computational overhead.

\par\medskip
\subsection{Total Loss}
\textbf{Total Objective.}  
Our overall training objective combines the base model detection loss with the two auxiliary terms, RCS and CMG.  
Formally, the total loss is defined as
\begin{equation}
\label{eq:total}
\mathcal{L}_{\mathrm{total}}
= \mathcal{L}_{\mathrm{det}}
+ \lambda_{\mathrm{rcs}}\,\mathcal{L}_{\mathrm{rcs}}
+ \lambda_{\mathrm{cms}}\,\mathcal{L}_{\mathrm{cms}} 
\end{equation}
where $\mathcal{L}_{\mathrm{det}}$ denotes the standard YOLOX \cite{ge2021yolox} detection loss(classification, regression, and objectness),  
$\mathcal{L}_{\mathrm{scs}}$ is the RoI-level supervised contrastive loss, and  
$\mathcal{L}_{\mathrm{cms}}$ is the feature-level cross-modal supervision loss.  
The balancing weights $\lambda_{\mathrm{scs}}$ and $\lambda_{\mathrm{cms}}$ control the relative contribution of each auxiliary objective.

\par\medskip
\section{EXPERIMENT}

\subsection{Experiment Setting}
\textbf{Datasets}.
The FLIR dataset \cite{flir} was collected for autonomous driving and ADAS research, providing temporally aligned RGB–thermal image pairs with COCO-style bounding box annotations. Although the dataset includes 15 object categories, it is common practice to evaluate only three: person, bicycle, and car. The raw distribution consists of 9,233 RGB and 9,711 thermal images. However, due to pixel-level misalignment between the two modalities, we adopt the widely used FLIR-aligned subset, which was manually corrected to improve registration quality. FLIR-aligned contains 5,142 pairs, split into 4,129 training pairs and 1,013 test pairs, and has become the de facto benchmark for evaluation.


\textbf{Evaluation Metric}.
We report results using mean Average Precision(AP), including mAP,  \( mAP_{50} \), \( mAP_{75} \), \( mAP_{S} \), \( mAP_{M} \), \( mAP_{L} \). On FLIR-aligned, evaluation is performed over the three classes (person, bicycle, car).

\begin{table}[t]
  \centering
  \caption{Quantitative results on FLIR dataset.}
  \label{tab:flir}
  \scriptsize
  \setlength{\tabcolsep}{3pt}
  \renewcommand{\arraystretch}{1.05}
  \resizebox{\columnwidth}{!}{%
  \begin{tabular}{@{}c|c|c|c|c|c|c@{}} 
    \hline
    
    Model & Data & Backbone & mAP$_{50}$ & mAP$_{75}$ & mAP & mAR \\
    \hline
    \multicolumn{7}{c}{{Comparison with Mono-Modality (Thermal) Methods}}\\
    \hline
    Faster R-CNN \cite{girshick2015fast-rcnn} & T & ResNet50     & 74.4 & 32.5 & 37.6 & 49.7 \\
    SSD \cite{liu2016ssd}          & T & VGG16        & 65.5 & 22.4 & 29.6 & 44.3 \\
    RetinaNet \cite{lin2017retinanet}    & T & ResNet50     & 64.5 & 20.3 & 28.3 & 44.4 \\
    YOLOv3 \cite{redmon2018yolov3}       & T & Darknet53    & 73.6 & 31.3 & 36.8 & 46.5 \\
    YOLOv5 \cite{jocher2021yolov5}       & T & CSPD53       & 73.9 & 35.7 & 39.5 & 47.3 \\
    YOLOF \cite{chen2021yolof}         & T & ResNet50     & 74.9 & 26.7 & 34.6 & 47.9 \\
    DDOD \cite{chen2021DDOD}         & T & ResNet50     & 72.7 & 26.2 & 33.9 & 28.2 \\
    YOLOX-L \cite{ge2021yolox}      & T & CSPD53       & 80.9 & 37.5 & 42.0 & 52.2 \\
    YOLOv7 \cite{wang2023yolov7}       & T & E-ELAN       & 75.6 & 32.2 & 38.2 & 49.0 \\
    TIRDet \cite{wang2023tirdet}         & T & CSPD53       & 81.4 & 41.1 & 44.3 & 54.0 \\
    \textbf{CGDet (ours)} & T & MobileNetV2 &
    \textbf{82.8} & \textbf{47.2} & \textbf{47.1} & \textbf{54.6}    \\
    \hline
    \multicolumn{7}{c}{{Comparison with Multispectral (Visible + Thermal) Methods}}\\
    \hline
    CFR\_3 \cite{zhang2020crf_3}      & V+T & VGG16   & 72.4 & --   & --   & --   \\
    GAFF \cite{zhang2021gaff}         & V+T & ResNet18& 72.9 & 32.9 & 37.5 & -- \\
    GAFF \cite{zhang2021gaff}        & V+T & VGG16   & 72.7 & 30.9 & 37.3 & -- \\
    YOLOFusion \cite{qingyun2022yolofusion}   & V+T & VGG16   & 76.6 & --   & 39.8 & -- \\
    CFT \cite{qingyun2021cft}          & V+T & CFB     & 78.7 & 35.5 & 40.2 & 52.5 \\
    InfusionNet \cite{yun2022InfusionNet}  & V+T & Infusion& 79.1 & 35.2 & 40.3 & -- \\
    CMX \cite{zhang2023cmx}          & V+T & Swin-T  & 82.2 & 37.1 & 42.3 & --   \\
    IGT \cite{chen2023igt}          & V+T & Swin-T  & \textbf{85.0} & 36.9 & 43.6 & --   \\
    TIRDet \cite{wang2023tirdet}         & T   & CSPD53  & 81.4 & 41.1 & 44.3 & 54.0 \\
    \textbf{CGDet (ours)} & T & MobileNetV2 &
    {82.8} & \textbf{47.2} & \textbf{47.1} & \textbf{54.6}    \\
    \hline
  \end{tabular}
  }
\end{table}

\textbf{Training Details}.
The training process is conducted for 100 epochs with a batch size of 4 on a single NVIDIA A5000 GPU. We use the SGD optimizer with momentum. The initial learning rate is set to 1e-2. Experiments are performed on the FLIR dataset.
A thermal image is first converted to a visible counterpart by Pearl-GAN \cite{luo2022pearl-gan}. The thermal and synthesized visible images are fused and passed through the MobileNetV2 backbone \cite{sandler2018mobilenetv2} and the lightweight FPN. The detector head operates on features refined by our two modules: RCS and CMG. In RCS, the RoI grid size is 5×5. In CMG, we set the minimum RoI side-length threshold to 3 and the RoIAlign \cite{he2017mask-rcnn} output resolution to 7. The final predictions are produced by the detection head.

\par\medskip
\subsection{Main Results}
\textbf{Quantitative Results on the FLIR Dataset.}

Table 1 summarizes object detection performance on the FLIR dataset. Despite using a MobileNetV2 backbone \cite{sandler2018mobilenetv2}, our CGDet achieves the best results with \( {mAP_{50}} \) 82.8, \( {mAP_{75}} \) 47.2, mAP 47.1, and mAR 54.6. This yields meaningful improvements in mAP: 2.8 over TIRDet and 5.1 over YOLOX-L \cite{ge2021yolox}. CGDet matches TIRDet at \( {mAP_{50}} \) and attains the highest \( {mAP_{75}} \). Furthermore, when compared with multispectral(visible+thermal) methods, the mono-modality CGDet surpasses most entries—including IGT \cite{chen2023igt}, CFT \cite{qingyun2021cft}, CMX \cite{zhang2023cmx}—demonstrating competitive performance without additional modalities. In summary, the proposed approach delivers consistent quantitative gains over prior mono-modality, multispectral base models while relying only on a lightweight backbone and a mono-modality input.

\begin{table}[t]
\caption{ Comparison of parameters and GFLOPs between TIRDet-S and Ours on the FLIR dataset.}
\centering
\setlength{\tabcolsep}{6pt}
\begin{tabular}{@{}l|c|c|c|c@{}}
\toprule
Method & mAP & mAR & Params(M) & FLOPs(G) \\
\midrule
TIRDet-S \cite{wang2023tirdet}  & 41.7 & 51.4 & 9.53 & 339.7 \\
Ours      & \textbf{47.1} & \textbf{54.6} & \textbf{3.15} & \textbf{331.4} \\
\midrule
\end{tabular}
\end{table}

\par\medskip
\begin{figure}[t]
  \centering
  \includegraphics[width=\linewidth,height=.88\textheight,keepaspectratio]{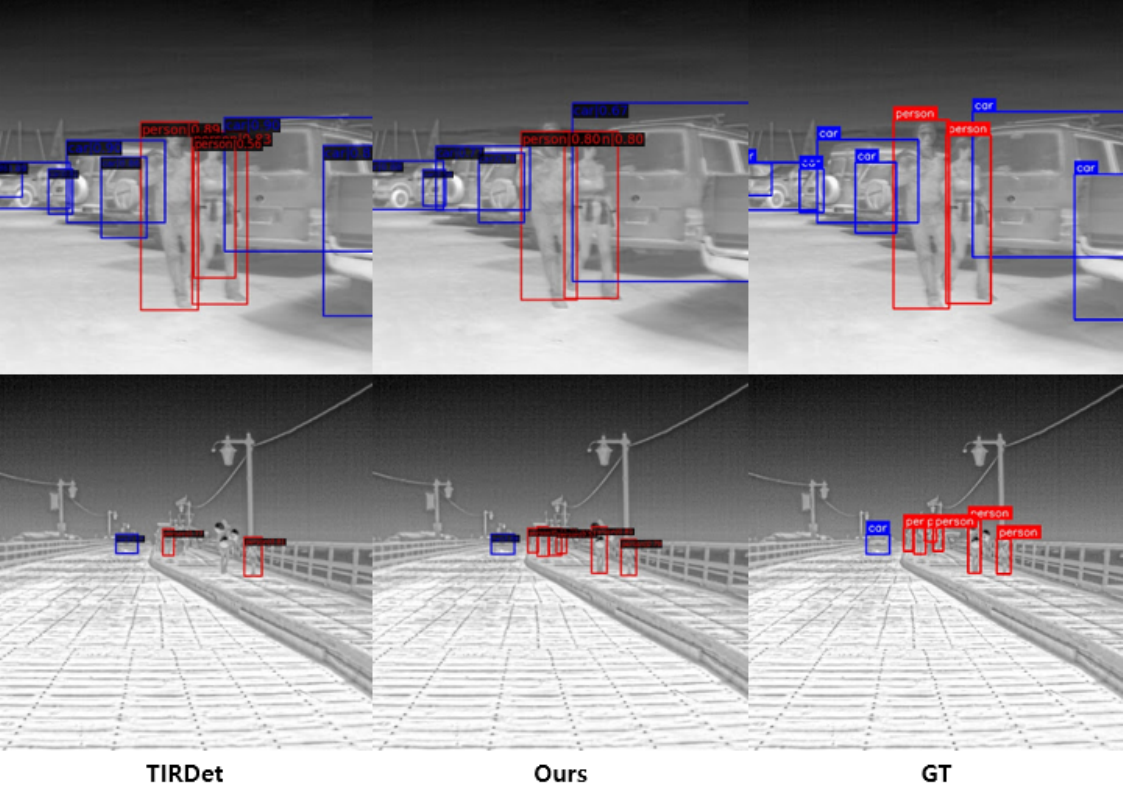}
  \vspace{-0.7cm}
  \caption{Qualitative comparison on FLIR dataset. Each triplet shows TIRDet(left), Ours(middle), and GT(right). Red, green, and blue boxes denote person, bike, and car, respectively.}
  \vspace{-2mm}
  \label{fig:tirdet}
\end{figure}

\begin{figure}[t]
  \centering
  \includegraphics[width=\linewidth,height=.88\textheight,keepaspectratio]{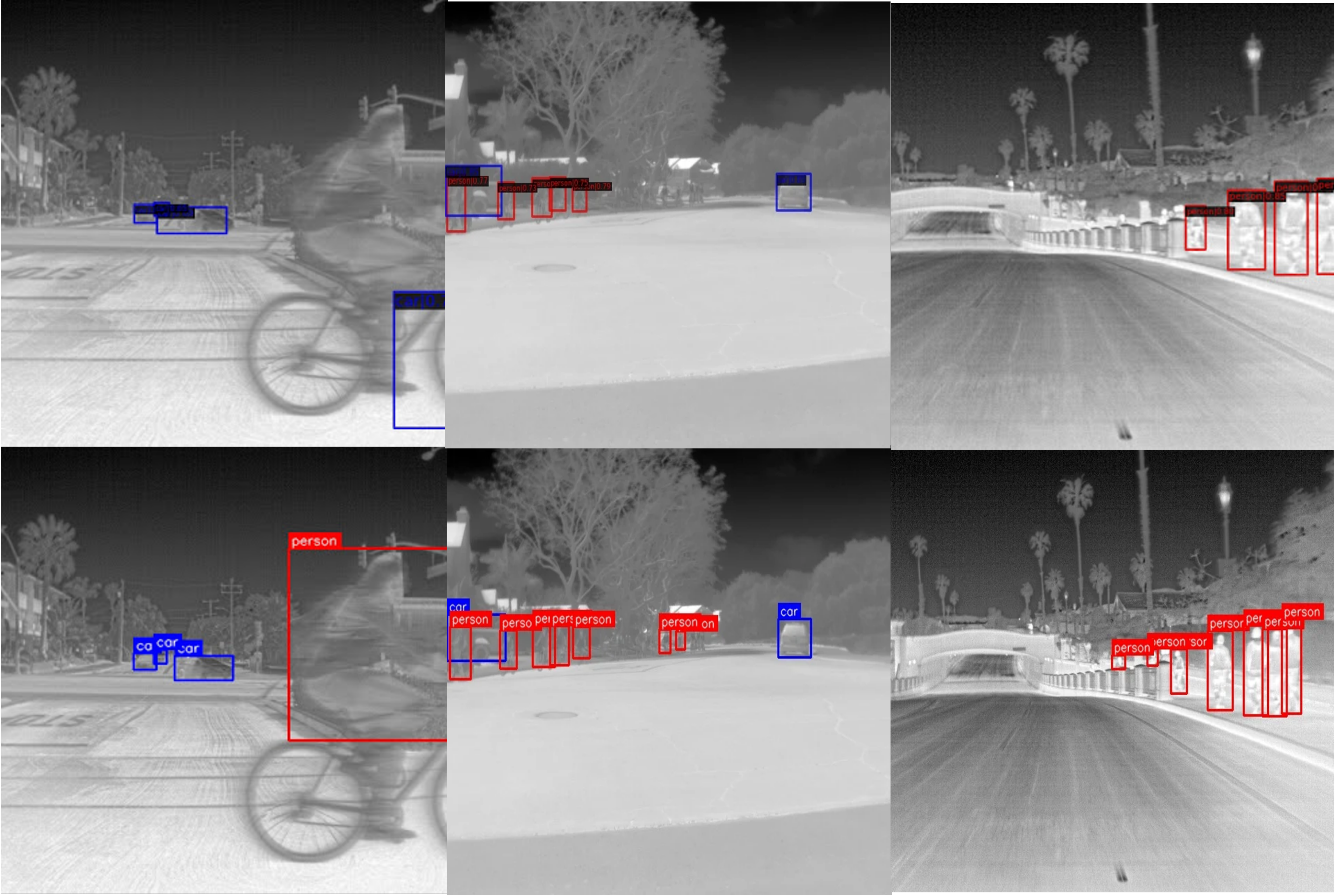}
  \vspace{-0.7cm}
  \caption{Failure case on FLIR dataset. Top: Ours. Bottom: GT. Columns—left: motion blur and weak edges lead to a missed cyclist; middle, right: distant, small pedestrians are not detected.}
  \vspace{-2mm}
  \label{fig:tirdet}
\end{figure}



\textbf{Qualitative comparison}.
Fig. 3 shows qualitative comparisons on the FLIR dataset. In row 1, TIRDet produces redundant, overlapping detections for the same class, whereas CGDet yields non-overlapping boxes that closely align with the ground-truth extents. In row 2, TIRDet fails to detect several pedestrians, while CGDet successfully recovers these missed targets. Using the same score threshold and NMS, these results indicate that CGDet improves both localization and recall, demonstrating the effectiveness of the proposed RCS and CMG modules.

\textbf{Accuracy–Efficiency Comparison on FLIR dataset}
Table II compares CGDet with TIRDet-S on the FLIR dataset in terms of both accuracy(mAP/mAR) and efficiency(Params, FLOPs). CGDet attains 47.1 mAP and 54.6 mAR, improving over TIRDet-S by +5.4 mAP and +3.2 mAR while using only 3.15M parameters—66.9\% fewer than TIRDet-S (9.53M)—and slightly fewer computations. All efficiency numbers are measured on the detector-only path, with training-time objectives disabled at inference. These results indicate that CGDet delivers a strictly better accuracy–efficiency trade-off than TIRDet-S while preserving mono-modality inference.

\textbf{Failure case on FLIR dataset}
We analyze failure cases in Fig. 4. In the top-left panel, motion blur from a moving object results in a missed detection; in the middle and top-right panels, distant objects are not detected. This reflects an intrinsic weakness of thermal imagery—limited edge and fine-texture cues—making detection of dynamic and far objects challenging. While the proposed training objectives generally increase feature separability and suppress overlaps, their effect diminishes when same-class RoIs are scarce or when the RGB teacher is uncertain/mismatched. These observations suggest remedies such as strengthening the small-object scale via pyramid adjustments (e.g., adding a lower FPN level or relaxing the minimum RoI size) and adopting confidence-weighted teacher guidance during distillation.

\par\medskip
\section{CONCLUSIONS}

To address class confusion and the lack of semantic cues prevalent in low light settings, this paper proposes CGDet, a lightweight detector that relies solely on a thermal input. The RCS module increases the global separability of instance-level representations to suppress false detections. In contrast, the CMG module injects multi-level semantic cues from an RGB-trained teacher into the thermal student’s features. As a result, CGDet achieves state-of-the-art performance on FLIR dataset under the mono-modality setting and even surpasses several multispectral methods, pushing the upper bound of thermal detection.




\par\medskip




\bibliographystyle{ieeetr}
\bibliography{ref}

\end{document}